\documentclass[preprint,12pt]{elsarticle}

\usepackage{amssymb}

\usepackage{bm}
\usepackage{mathrsfs}

\journal{arxiv.org}

\begin{document}

\begin{frontmatter}



\title{Once for All: a Two-flow Convolutional Neural Network for Visual Tracking}


\author{Kai Chen \quad Wenbing Tao}

\address{\{chkap,wenbingtao\}@hust.edu.cn}

\begin{abstract}
One of the main challenges of visual object tracking comes from the arbitrary appearance of objects. Most existing algorithms try to resolve this problem as an object-specific task, i.e., the model is trained to regenerate or classify a specific object. As a result, the model need to be initialized and retrained for different objects. In this paper, we propose a more generic approach utilizing a novel two-flow convolutional neural network (named YCNN). The YCNN takes two inputs (one is object image patch, the other is search image patch), then outputs a response map which predicts how likely the object appears in a specific location. Unlike those object-specific approach, the YCNN is trained to measure the similarity between two image patches. Thus it will not be confined to any specific object. Furthermore the network can be end-to-end trained to extract both shallow and deep convolutional features which are dedicated for visual tracking. And once properly trained, the YCNN can be applied to track all kinds of objects without further training and updating. Benefiting from the once-for-all model, our algorithm is able to run at a very high speed of 45 frames-per-second. The experiments on 51 sequences also show that our algorithm achieves an outstanding performance.

\end{abstract}


\end{frontmatter}


\section{Introduction}
Visual object tracking has play an important role in numerical applications such as automated surveillance, traffic monitoring and unmanned aerial vehicle (UAV)~\cite{track_survey_2006}. Visual object tracking is challenging as the object is unknown before tracking and has an arbitrary appearance. As a result, most existing trackers, with either generative model or discriminative model, are all based on object-specific approach. In a generative model, the basis vectors used to represent an object need to be initialized when given a new object. Similarly, the discriminative classifier for object detection in a discriminative model need to be retrained when tracking in a new sequences. Specifically, the $l_1$ tracker~\cite{L1T} tries to represent an object by target templates and trivial templates, however those templates are learned from the given object in first frame. The recent KCF~\cite{KCF} tracker uses a kernelized ridge regression model, which needs to be trained and updated through frame to frame, to predict the object location. Though these object-specific trackers have demonstrated outstanding robustness and accuracy, two natural defects need to be overcame. First, the tracking procedure tends to be time-consuming because of the frequent training and updating. Second, the tracker is more likely to drift away from object especially during a long-term tracking, which also resulting from frequent updating.

Recently, convolutional neural networks (CNN) have shown a great success in a number of computer vision tasks such as image classification, object detection, face recognition and so on. However, the representation power of CNN seems not suited to visual tracking as the object varies from sequence to sequence and only one object instance is provided before tracking. It will be a tricky work to train a proper CNN in an object-specific approach. One alternative solution is to transfer the CNN pretrained from large scale image classification datasets like ImageNet~\cite{ILSVRC15}. But this will significantly weaken the power of CNN because of the huge gap between classifying an object and predicting the location of an object.

In this paper, we propose an object-free approach to predict the object location. Unlike the usual convolutional neural networks, which only one image is passed through the convolutional layers, here we take two convolutional flows, one is for the object patch to be tracked, the other is for the search patch where the object may appear. What's more, in each flow both shallow and deep convolutional features are adopted as the shallow features are useful to discriminate the object from background and the deep features show superiority of recognising an object with varying appearance. Then all of the two-flow features are concatenated and passed through the fully connective layers to output a two dimensional prediction map, which shows where and how likely the object is to appear in the search patch. Due to lack of available labeled tracking sequences, our YCNN is firstly trained with search patches and object patches clipped from images in ImageNet~\cite{ILSVRC15}. To simulate the case of tracking real object with varying appearance, those object and search patches are manually manipulated by rotating, translating, adding noise and so on. Finally the YCNN is fine-tuned with data pairs retrieved from labeled video sequences.

In a typical tracking task, there may exist various challenges such as deformation, partial occlusion and rotation. To handle these challenges, we also propose a confidence-based tracking framework. To each tracked object patch, we assign an confidence score based on how well the object was tracked. When predicting in a new frame, a number of tracked object patches are selected to predict object location, but each of them is weighted by the corresponding object confidence score. Then the final location can be predicted by the weighted mean maps. With such a framework, our tracker will be robust to occluded objects and also be adapted to deformation or rotation.

Compared to most object-specific approaches, our YCNN based tracker has three main features. First, this is, as far as we known, the first once-for-all approach, \textit{i.e.} once trained, ready to track all. Furthermore it can run at a high speed as no online training needed. Second, the YCNN is compact and can be trained end-to-end, in which the power of CNN can be fully exploited. The last one, as an unexpected benefit, is that, the YCNN is trained to predict more likely an object rather than background. And thus it will be robust to the spatial perturbation of object patch.

\section{Related Work}
Most visual tracking algorithms are based on either generative model or discriminative model. In generative models, a valid object candidate is supposed to be reconstructed with a number of templates learned from the initial object. For example, Ross \textit{et al.}~\cite{IVT} proposed a subspace model, based on incremental algorithms for principal component analysis, to represent the object appearance. Also, sparse coding~\cite{L1T,RT_L1T} can be exploited to reconstruct the target. Another approach, as the discriminative models usually do, is to develop a classifier which discriminating the object from background. A number of discriminating trackers incorporating various models such as boosting~\cite{Boosting}, multiple instance learning~\cite{MIL}, structured SVM~\cite{STRUCK}, and kernelized correlation filter~\cite{KCF} have achieved great success. However the above mentioned trackers are all limited to hand-crafted features and need to be retrained and updated frequently.

The main challenge of applying deep convolutional neural networks to visual tracking is that the available labeled tracking sequences are far from enough to train a CNN based classifier for a specific object. Thus, most existing method try to transfer a CNN pretrained for image recognition such as VGG-Net~\cite{deep_cnn}. In~\cite{HCF_CF}, a pretrained convolutional networks is used to extract both shallow and deep convolutional features, then those features are utilized to predict the target location with correlation filters. L. Wang \textit{et al.}~\cite{FCN_GS} proposed a general network to capture the category information of target and a specific network to discriminate the object from background. In~\cite{DSM}, CNN is adopted to predict a target-specific saliency map which highlights the regions discriminating target from background. Note that, the basic CNN features used in~\cite{HCF_CF,FCN_GS,DSM} are all originally trained for image recognition, which may not fit the visual tracking task. Recently, H. Nam \textit{et al.}~\cite{MDCNN} proposed a multi-domain convolutional neural network for visual tracking which is composed of shared layers and multiple branches of domain-specific layers. In such a way, the network is able to be fully pre-trained with video sequences. However, those object-specific approaches~\cite{FCN_GS,DSM,MDCNN} all need to update the model online, and run at a relatively low speed.

%

\begin{figure}[t]
  \centering
  \includegraphics[width=1.0\linewidth]{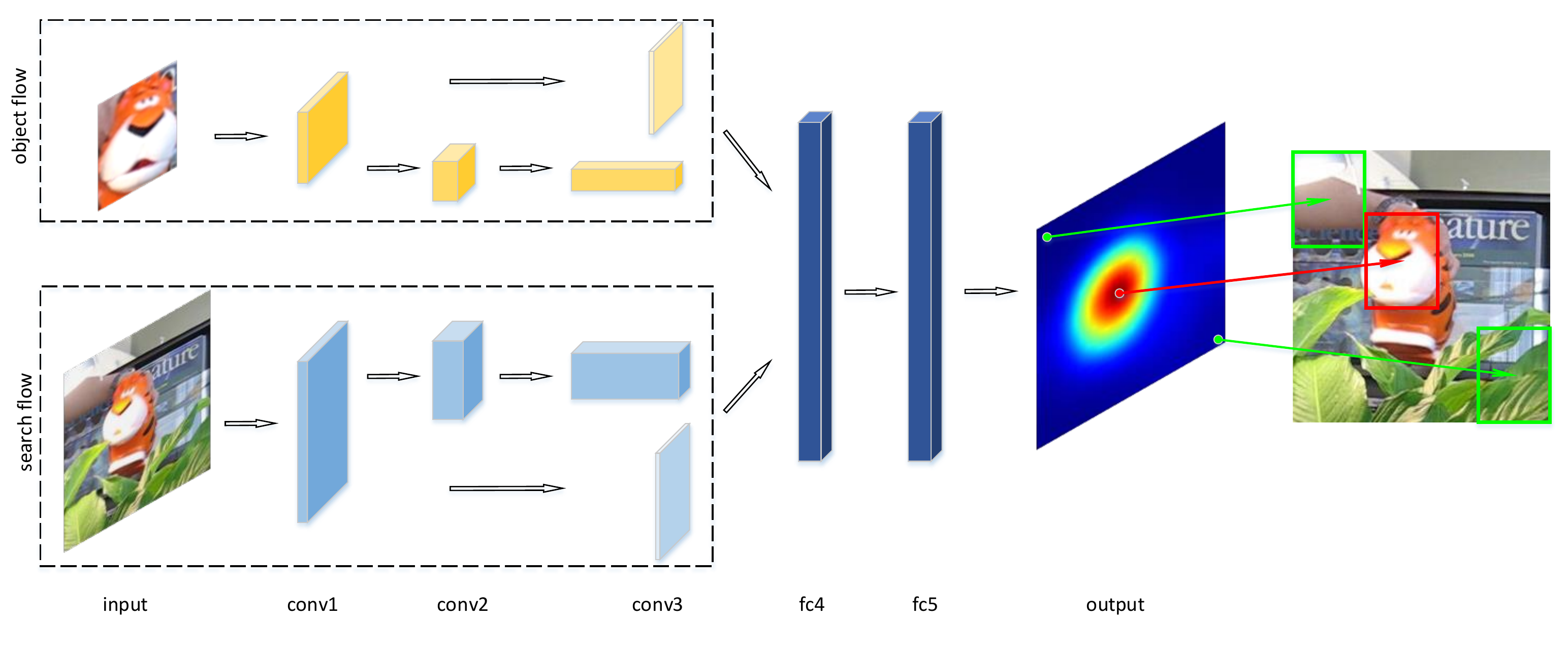}
  \caption{Schematic diagram of YCNN.}
  \label{fig:architecture}
\end{figure}

\begin{table}[t]
  \caption{Architecture of YCNN. The architecture consists of 3 convolutional layers and 3 fully connective layers. The column of `conca.' indicates a layer that concatenates all the convolutional features generated in layer of `conv3'. Each convolutional layer is denoted as `$num\times size \times size\quad \textrm{st.}\ s\quad \textrm{pool}\ p$', where \textit{num} means the number of convolutional filters and \textit{size} means receptive size of the filters. $s$ and $p$ indicate the convolutional stride and the max-pooling downsampling factor respectively. The RELU~\cite{RELU} activation function is used in all layers excluding the `output' layer.}
  \label{tab:architecture}
  \centering
  \includegraphics[width=1.0\linewidth]{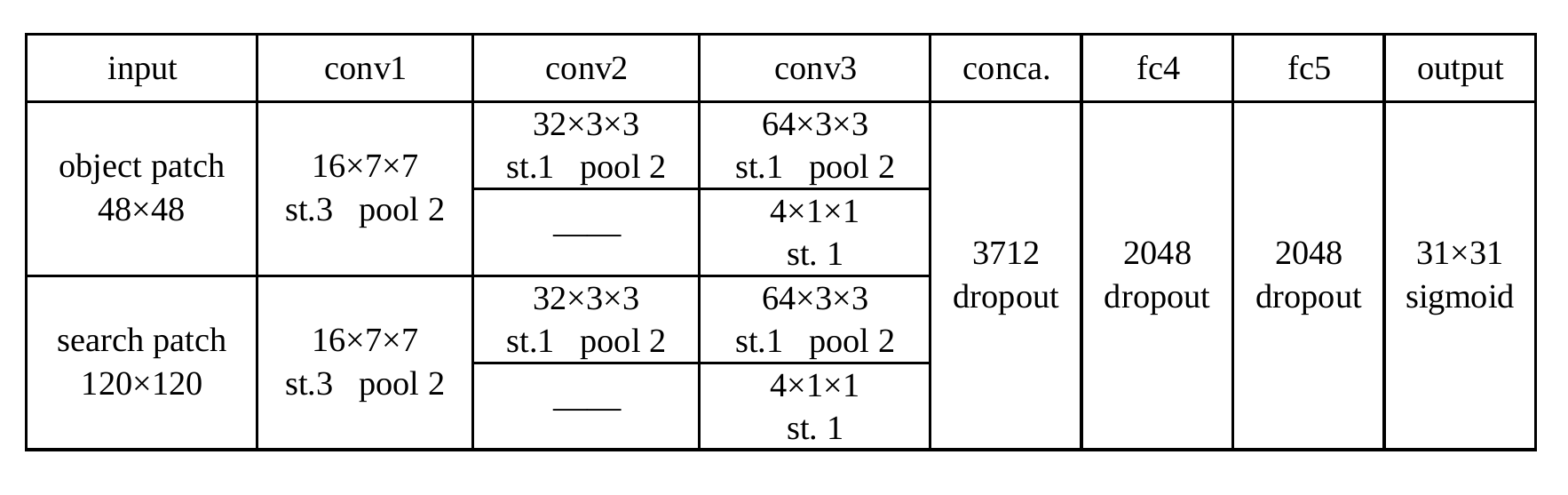}
\end{table}

\section{YCNN}
The basic motivation of the proposed YCNN comes from the idea that, instead of tracking an object by classifying a candidate, we can design a classifier to judge how the candidate looks like the object. In this way, the classifier will not be limited to a specific object, which is perfect when designing a CNN based classifier. Technically it is practicable to develop a convolutional networks which take two images with same sizes and output a scalar value which measures how the two images look like each other. But it will be complicate and redundant when tracking an object as there will be lots of candidates to be compared with the target. A more intelligent and efficient approach is to develop a CNN which takes an object image and a search image, which is much larger than the object image, and outputs a prediction map which indicates how likely the object is to appear in the search image.

\subsection{Architecture}

The schematic diagram of our YCNN is shown in figure~\ref{fig:architecture}. We have noticed that a much deeper network such as the VGG-Net~\cite{deep_cnn} with 5 convolutional layers is powerful to capture the sematic information for object detection or image recognition. But in our model, the main task is to measure the similarities between two images, and deep sematic information will be redundant and expensive for this task. Here, we build a three-layer hierarchic convolutional network to extract both shallow and deep features. The shallow features can be extracted from the first convolutional layer. To reduce the dimensions of shallow features, a convolutional layer with only 4 filters is appended. The more detailed network settings is listed in table~\ref{tab:architecture}. Note that, both the object flow and search flow share the same convolutional filters so as to reduce the number of parameters to be trained.

\subsection{Loss Function}

For training the YCNN, we assign a labeled prediction map to each pair of object image and search image. The labeled map $\bm{M}_\mathrm{L}$ follows an Gaussian shape, and the peak with value $1$ indicates the real location of the object in the search image. A straightforward way to define the loss function for YCNN can be like this,
\begin{equation} \label{equ:naive_loss}
\mathrm{L_0}(\bm{M},\bm{M}_\mathrm{L})=\left\| \bm{M}-\bm{M}_\mathrm{L} \right\|_2^2.
\end{equation}
Here, $\left\|\cdot\right\|_2$ means the $l\textrm{-}2$ norm. $\bm{M}$ denotes the output map of YCNN and each of the entries in $\bm{M}$ can be regarded as a prediction sample for the corresponding location.  This definition is quite simple and efficient for computing, but it does not work very well in practice. In fact, our initial attempting shows that, with such a kind of loss function the YCNN is to be stuck at a locally optimal point and tends to output a plain zero map. This predicament is supposed to be resulting from two issues. First, the tracking is based on positive predictions (\textit{i.e.} larger values in $\bm{M}$), and more attention should be paid to make positive predictions with less error. But in equation~\ref{equ:naive_loss}, both positive and negative predictions are evenly weighted. Second, as nearly 95 percent of entries in the training label $\bm{M}_\mathrm{L}$ will be near 0, the contribution of positive labels would be easily submerged. Thus the YCNN is probably to be trained to output a zero map. To deal with this predicament, we design a revised loss function as follows.

\begin{equation} \label{equ:weight_func}
\mathrm{W}(\bm{M}_\mathrm{L})=a\cdot\mathrm{exp}(b\cdot\bm{M}_\mathrm{L}),
\end{equation}
\begin{equation} \label{equ:sign_func}
\mathrm{S}(\bm{M},\bm{M}_\mathrm{L})=\frac{\mathrm{sign}(\left| \bm{M}-\bm{M}_\mathrm{L} \right|-Th)+1}{2},
\end{equation}
\begin{equation} \label{equ:final_loss}
\mathrm{L}(\bm{M},\bm{M}_\mathrm{L})=\left\| \mathrm{W}(\bm{M}_\mathrm{L}) \odot \mathrm{S}(\bm{M},\bm{M}_\mathrm{L}) \odot (\bm{M}-\bm{M}_\mathrm{L}) \right\|_2^2.
\end{equation}

Equation~\ref{equ:weight_func} defines a exponential weighting map, in which the losses for positive predictions will be highly weighted while for negative predictions strongly decayed. $a$ and $b$ here denote the factors to reshape the weighting map. The sign function $\mathrm{sign}(x)$ used in equation~\ref{equ:sign_func} returns $1$ if $x\ge0$ otherwise $-1$. So equation~\ref{equ:sign_func} defines a binary indicating map in which $1$ means the absolute error between prediction and label is greater than or equal to a given threshold $Th$ while 0 means less error. Finally the improved loss function is defined in equation~\ref{equ:final_loss}. Here $\odot$ means element-wise product. By masking the original error map with the indicating map, most of the negative samples would be significantly suppressed while the positive samples almost not influenced. This is because the prediction errors of negative samples tends to be small but with a large amount, while for positive samples they will be large but less amount. In our experiment, $a=0.1$, $b=3$ and $Th=0.05$.

\subsection{Two-stage Training}

How to generate enough data pairs to train such an CNN with more than 10 millions of parameters is another challenge. The training data of object patches and search patches can be extracted from different frames of a tracking sequences. But only hundreds of tracking sequences are publicly available and the object appearances in the tracking sequences are too monotonous to train the YCNN for general object tracking. To solve this problem, we try to firstly train the YCNN with single image in ImageNet~\cite{ILSVRC15}.

\begin{figure}[t]
  \centering
  \includegraphics[width=1.0\linewidth]{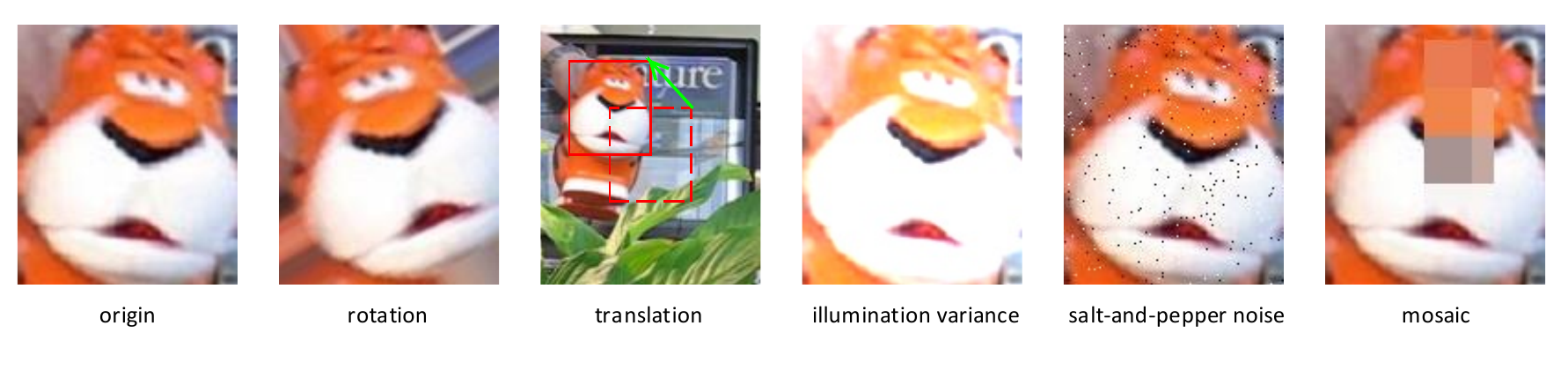}
  \caption{Examples of data augmentation. Rotation apply only to object patch, while translation only to search patch. All the rest apply to both object and search patches. The search patch is always translated randomly.}
  \label{fig:image_perturbation}
\end{figure}

\paragraph{Training with single image} \label{sec:training_with_image}
ImageNet has provided millions of high-quality images with numerous objects which is a perfect dataset for training a network with high generalization ability. Here we extract both object patch and search patch from a single image in ImageNet. In such a case, the object in search patch will be identical to the object patch, which is however not real when tracking in a video sequence. To simulate the real scenario, a number of data augmentation techniques, such as rotation, translation, illumination variation, mosaic, and salt-and-pepper noise, are adopted to manipulate both object patches and search patches as shown in figure~\ref{fig:image_perturbation}. Note that, the extracted training data are all limited to labeled objects in the image. And the YCNN, in some degree, is trained to predict the location of object other than background, \textit{i.e.} the objectness is taken into account. At this point, our YCNN will be more robust to spatial perturbation of the initial object. The Adam optimizer~\cite{Adam} is used to train the YCNN with learning rate of 1e-4. And the batch size is set to 256.

\paragraph{Fine-tuning with tracking sequence}

To make the YCNN more robust when tracking in real scenarios, we further fine-tune it with training data extracted from real tracking sequences. Both object patch and search patch can be clipped from different frames in a tracking sequences. It should be noted that the object patch and the object in the search patch should share a similar appearance and the object patch should appear before the search patch. Suppose the frame number for extracting object patch and search patch are $f_\mathrm{obj}$ and $f_\mathrm{sec}$ respectively. Then they must be subjected to $0< f_\mathrm{sec} - f_\mathrm{obj} \le \Delta f$. In our experiments, $\Delta f$ is set to 10 for elastic object otherwise set to 100 for rigid object. The object types (`elastic' or `rigid') in the training sequences are empirically labeled by human. Furthermore, those frames with heavily occluded object are also excluded. Unlike in the case of training from a single image, here  no data augmentation technique except translation is used. At this stage, the learning rate is reduced to 1e-5, and the batch size is set to 128.

\section{Visual Tracking via YCNN}

In a typical tracking sequence, the object appearance may undergo a significant change. To be adaptive to the change, $N$ previous tracked object patches are used to predict the location of object via YCNN. Those $N$ patches are randomly selected and each of them is assigned a confidence score which indicating how confident the object was tracked. The object patch with higher confidence score will have more weight when predicting the location whereas lower confidence means less weight.

Let $S_k$ be the search patch in $k$-th frame and $O_{a_i},i=1,2,\ldots,N$ be $N$ selected object patches with frame number $a_i$ respectively. And the prediction map outputted by YCNN can be defined as $\mathrm{Y}(O_{a_i},S_k)$ with given object patch $O_{a_i}$ and search patch $S_k$. Then the combined prediction map $\bm{M}_k$ and the prediction confidence score $c_k$ can be defined as follows.
\begin{equation} \label{equ:final_prediction}
\bm{M}_k=\frac{\sum_{i=1}^{N}{c_{a_i}\,\mathrm{Y}(O_{a_i},S_k)}}{\sum_{i=1}^{N}{c_{a_i}}} \quad\quad a_i < k, i=1,\ldots,N
\end{equation}
\begin{equation} \label{equ:confidence}
c_k=\max(\bm{M}_k)
\end{equation}

The location of object in $k$-th frame can be easily located with the index of the maximum value of $\bm{M}_k$. A wrongly tracked object may lead to drift in following frames. To get around this, we define a tracking confidence threshold $c_{\mathrm{Th}}$, and those object patches with confidence score less than $c_{\mathrm{Th}}$ will never be selected to predict the location.

For scale estimation, we use a naive implementation by repeating the above procedure with scaled search patches. In our experiments, we set $N$ to 5. And the confidence score $c_1$ for the initial given object patch is set as $\max(\mathrm{Y}(O_{1},S_1))$. The confidence threshold $c_{\mathrm{Th}}$ is then set to half of $c_1$.

\section{Experiments}

The experiments are conducted on the CVPR2013 benchmark~\cite{benchmark_2013}, which contains 51 frequently used tracking sequences. The initial training data are extracted from more than 1.2 million carefully labeled images provided by ImageNet~\cite{ILSVRC15}. The tracking sequences for fine-tuning the YCNN are collected from VOT2015~\cite{VOT2015} and TB-100~\cite{benchmark_2015}. Those sequences appeared in the above testing sequences are excluded.

\subsection{Overall Results}

The performance of a tracking algorithm is usually evaluated in two aspects. One is based on the Center Location Error and the other is based on the Overlap Rate, as in ~\cite{benchmark_2013}. For Center Location Error, the performance can be measured as Precision Plot which shows the percentage of frames whose estimated location is within the given threshold distance of the ground truth. The performance rank is based on the score of given threshold of 20 pixels. Similarly the performance can be evaluated by Success Plot based on Overlap Rate. For each given threshold for Overlap Rate, we can calculate the ratios of those frames whose overlap rate is over the threshold. Then the algorithms can be ranked according to the area under curve (AUC) of the Success Plot. To evaluate the robustness against both the spatial perturbation and temporal perturbation, the tracking algorithms are tested on spatial robustness evaluation (SRE) and temporal robustness evaluation (TRE), in addition to the usual one-pass evaluation (OPE). In SRE, the initial given boundary box is perturbed by shifting or scaling. In TRE, several segments of the original sequences are adopted to evaluate the performance.

\begin{figure}[t]
\centering
\begin{tabular}[t]{c c c c}
\includegraphics[width=0.3\linewidth]{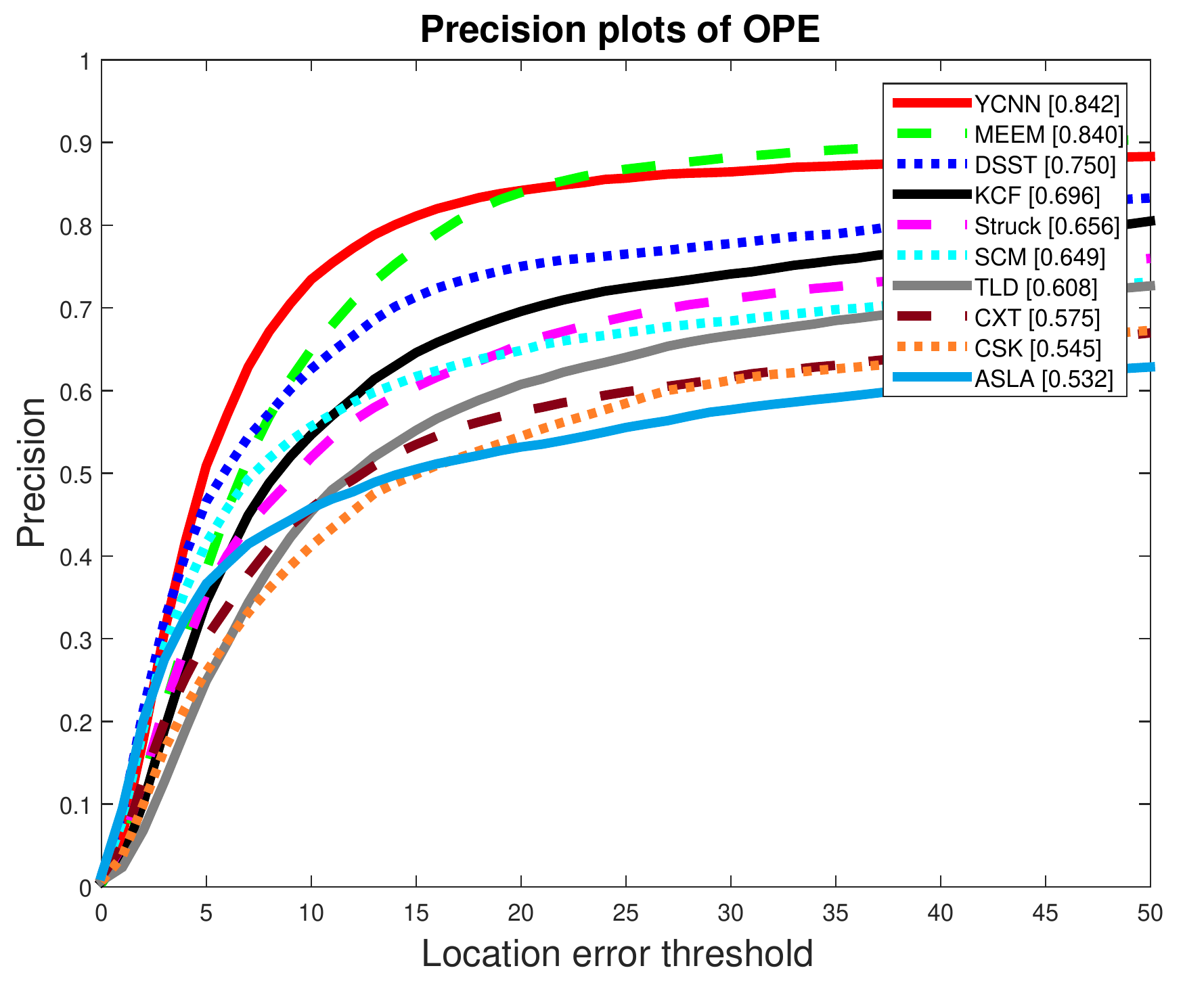} & \includegraphics[width=0.3\linewidth]{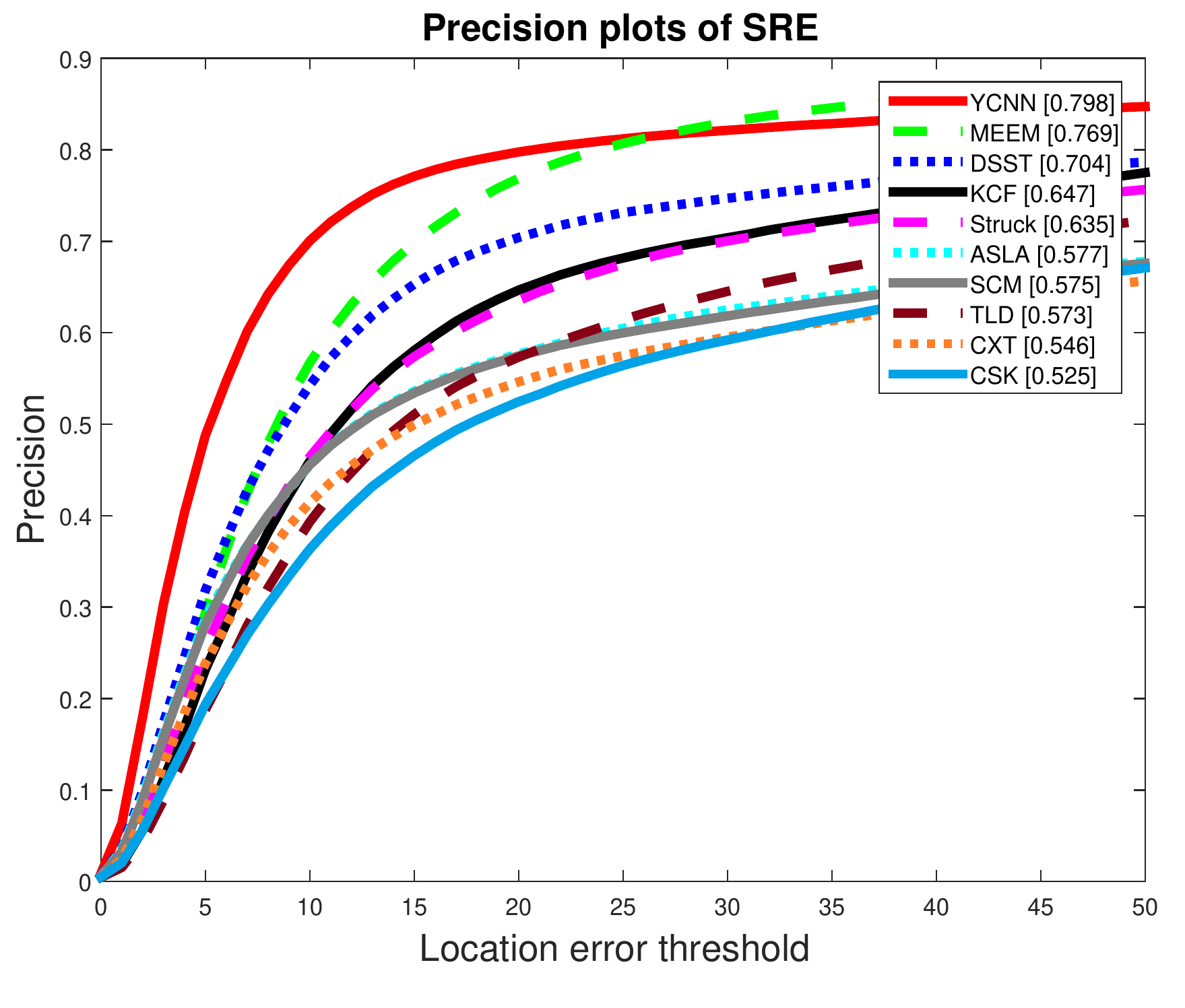} & \includegraphics[width=0.3\linewidth]{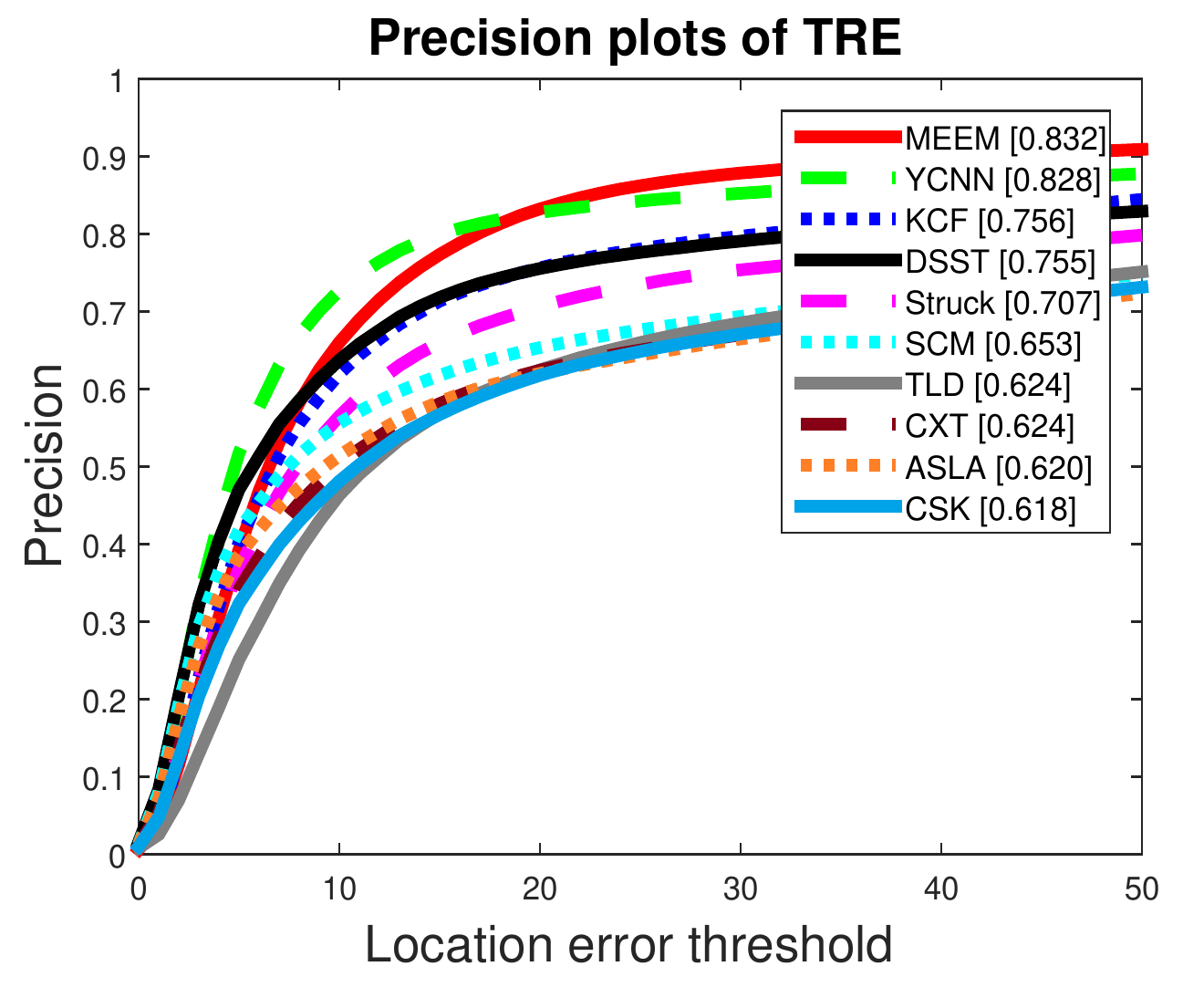} \\

\includegraphics[width=0.3\linewidth]{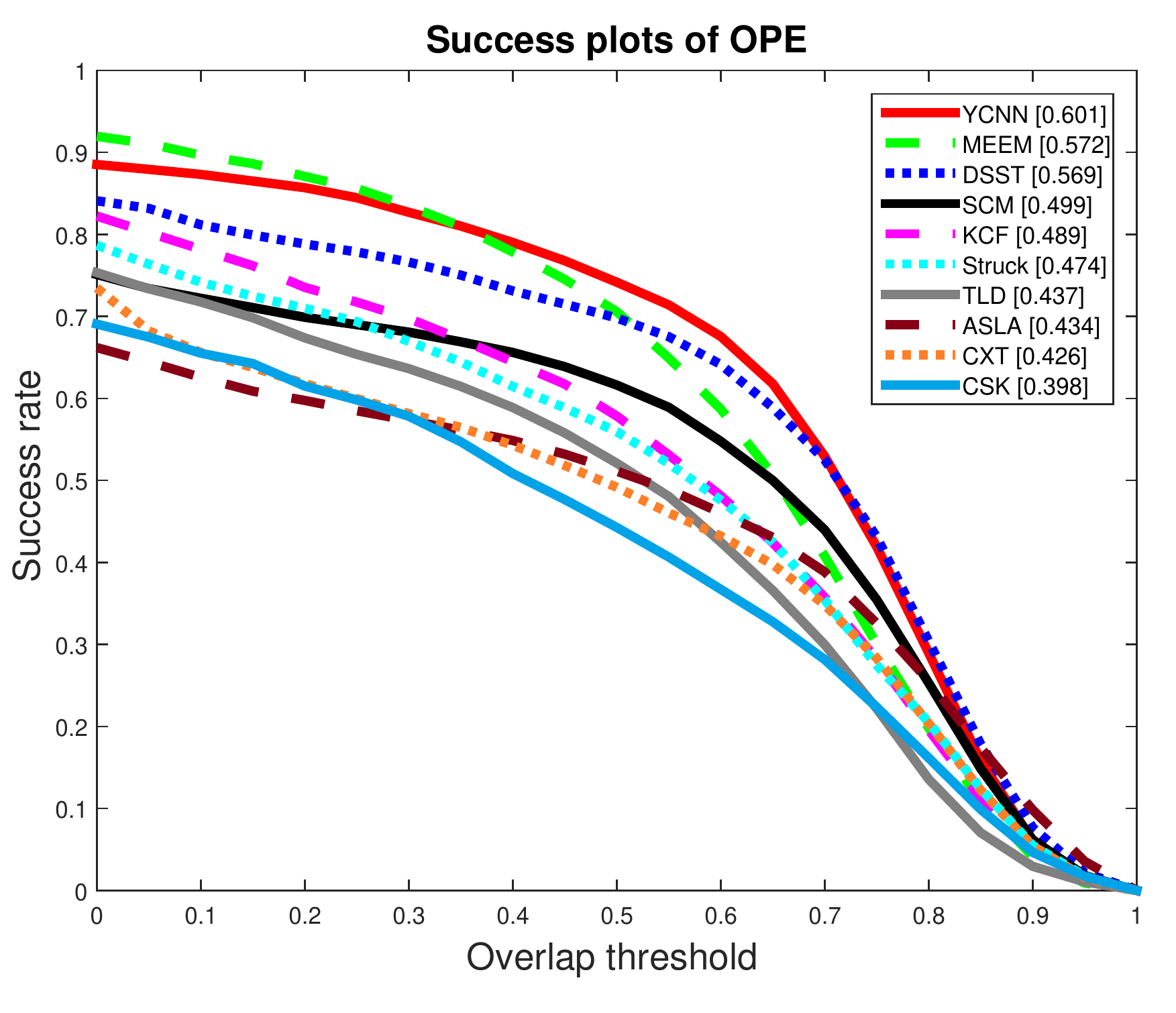} & \includegraphics[width=0.3\linewidth]{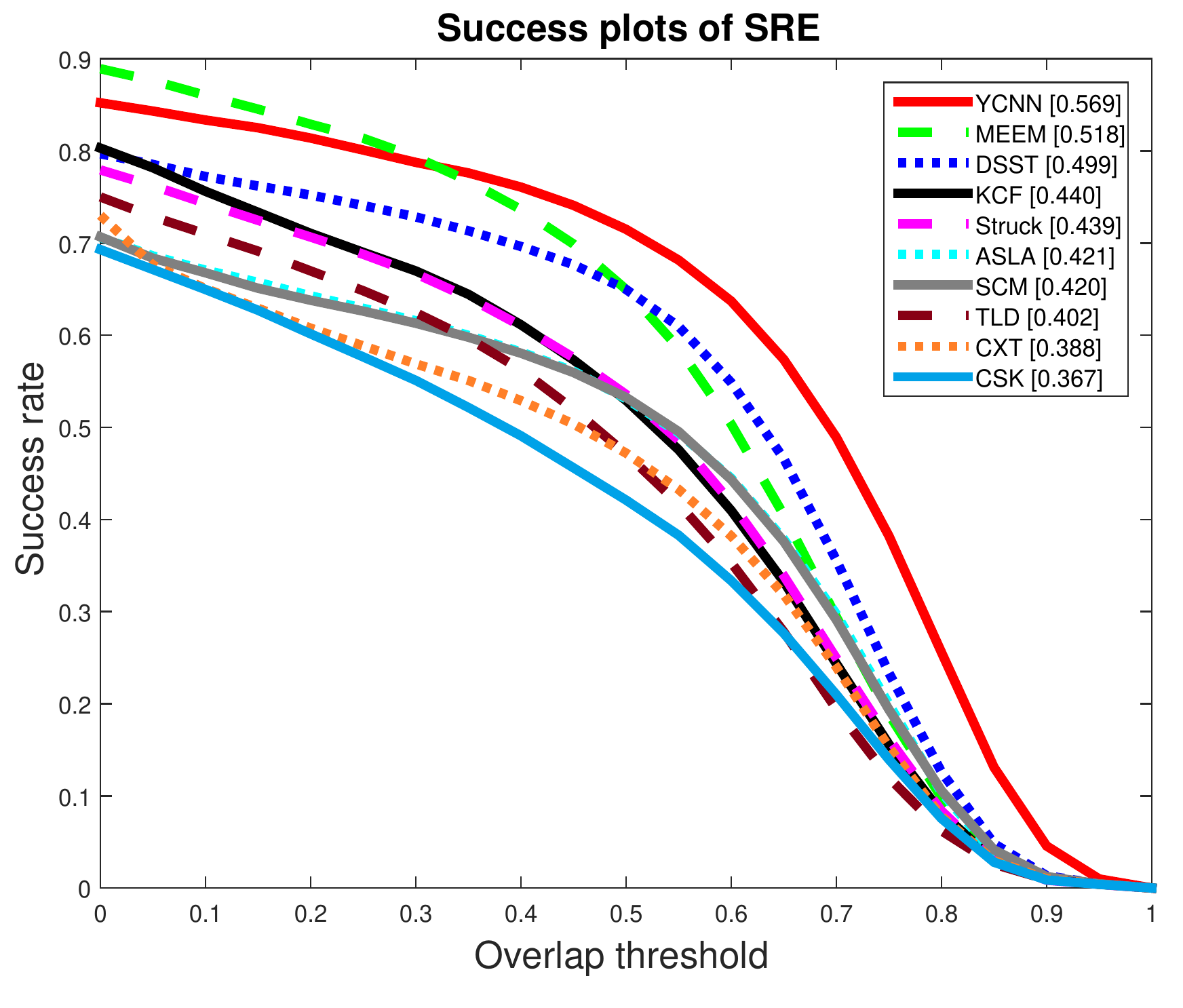} & \includegraphics[width=0.3\linewidth]{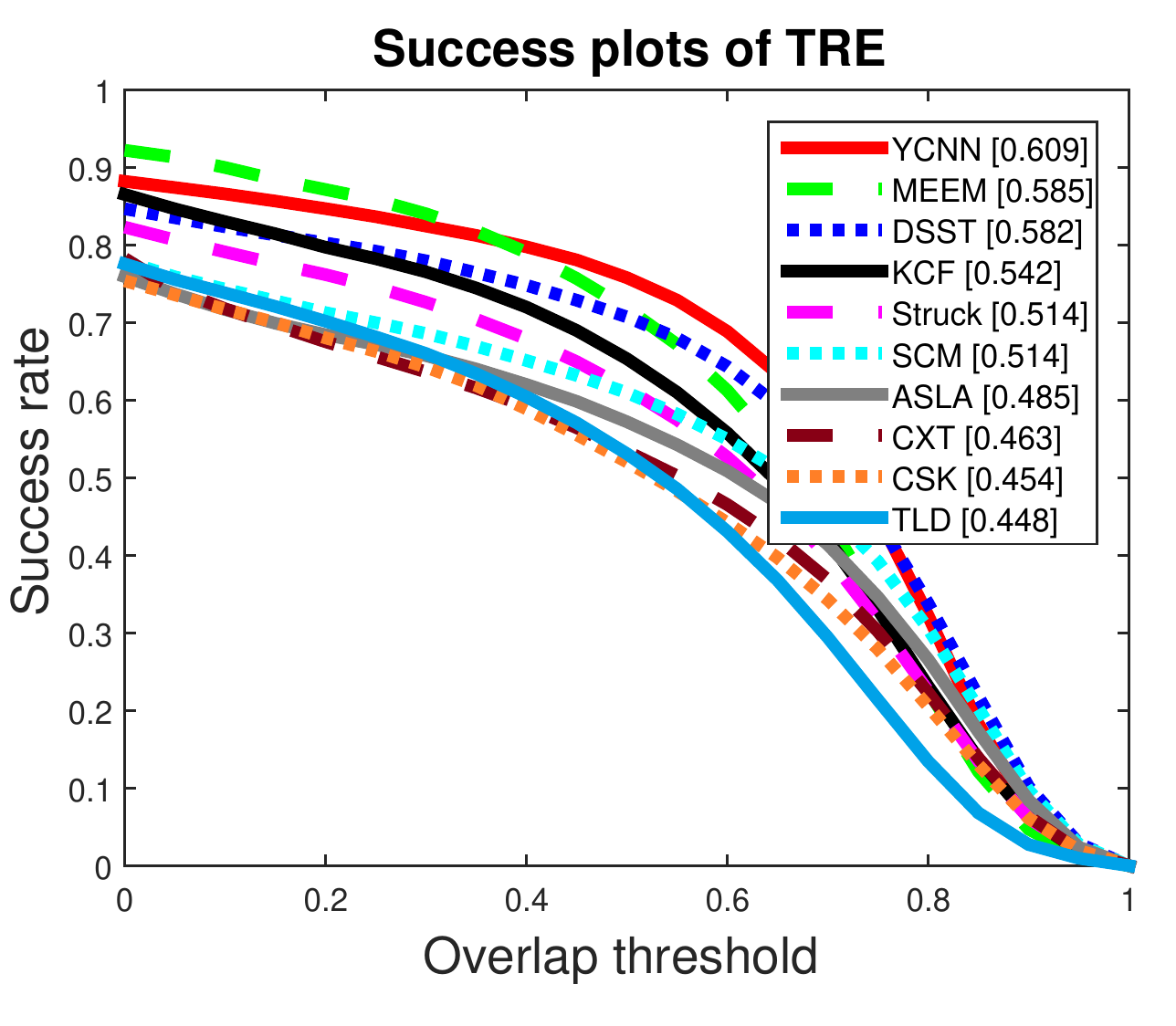} \\
\end{tabular}
\caption{Precision plots and success plots of YCNN and other 9 trackers in OPE, SRE, and TRE. In precision plots, the values in legend indicate the success rate with given threshold of 20 pixels. In success plots, the values means the area-under-curve score.}
\label{fig:overall_results}
\end{figure}

We compare our proposed algorithm (denoted as YCNN) with other 9 state-of-the-art tracking algorithms, such as Struck~\cite{STRUCK}, CSK~\cite{CSK}, TLD~\cite{TLD}, ASLA~\cite{ASLA}, SCM~\cite{SCM}, CXT~\cite{CXT}, KCF~\cite{KCF}, DSST~\cite{DSST}, MEEM~\cite{MEEM}. The overall results of Precision Plots and Success Plots in OPE, TRE, SRE are shown in figure~\ref{fig:overall_results}. The proposed YCNN outperforms the other 9 tracking algorithms in 5 of the 6 plots. The performance of YCNN only slightly outperforms the second ranked MEEM in OPE. But in SRE, the gap between YCNN and MEEM is enlarged. Especially in the success plots of SRE, the YCNN achieves an area under curve score of 0.569 which is 10\% more than MEEM. The drop in SRE is comprehensible as the spatial perturbation lead to higher probability of drifting from object. But the proposed YCNN is more robust against the spatial perturbation, which may be contributed to the end-to-end training for predicting the location of object rather than background, as mentioned in section \ref{sec:training_with_image}.

\begin{table}[t]
  \caption{Average precision scores based on center location error with given threshold of 20 pixels. The first and second best results are in red and green respectively.}
  \label{tab:precision_attribute_analysis}
  \centering
  \includegraphics[width=1.0\linewidth]{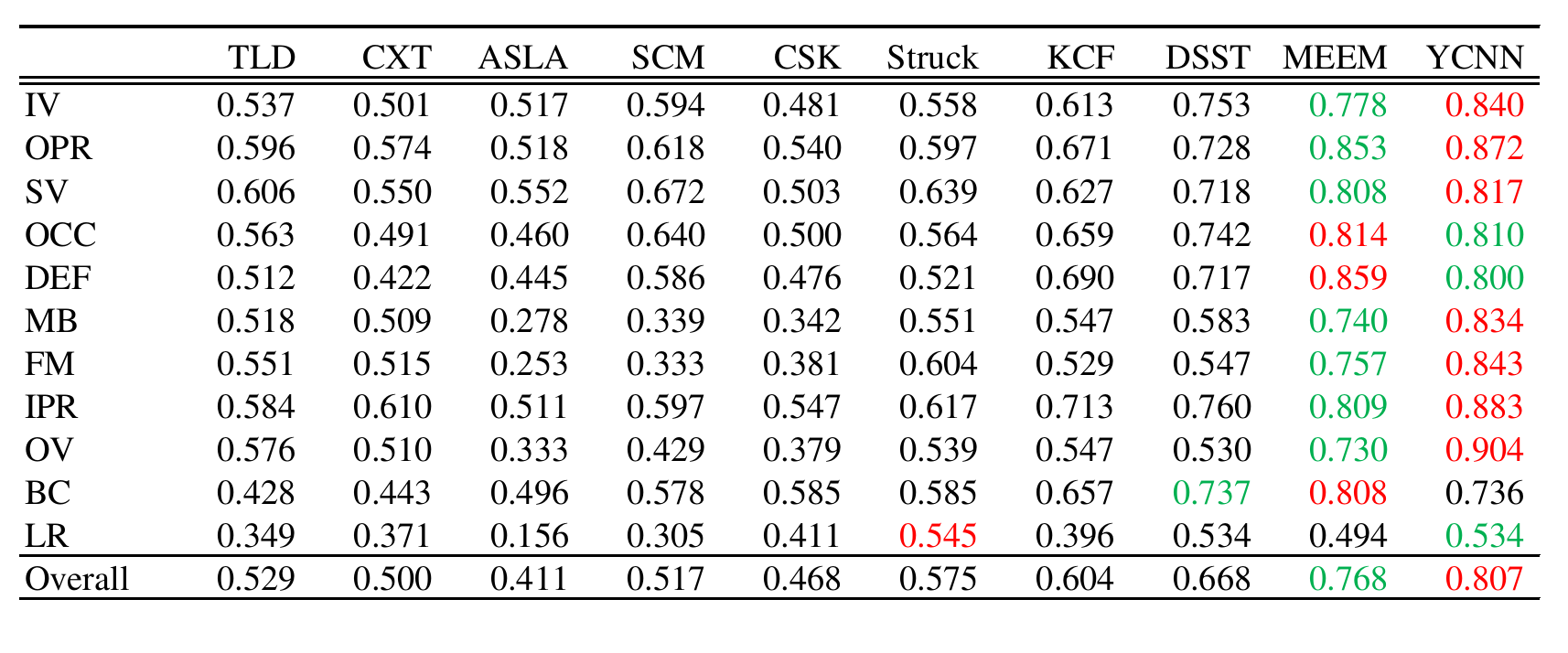}
\end{table}

\begin{table}[t]
  \caption{Average area-under-curve scores of success plot based on overlap rate. The first and second best results are in red and green respectively.}
  \label{tab:sucess_attribute_analysis}
  \centering
  \includegraphics[width=1.0\linewidth]{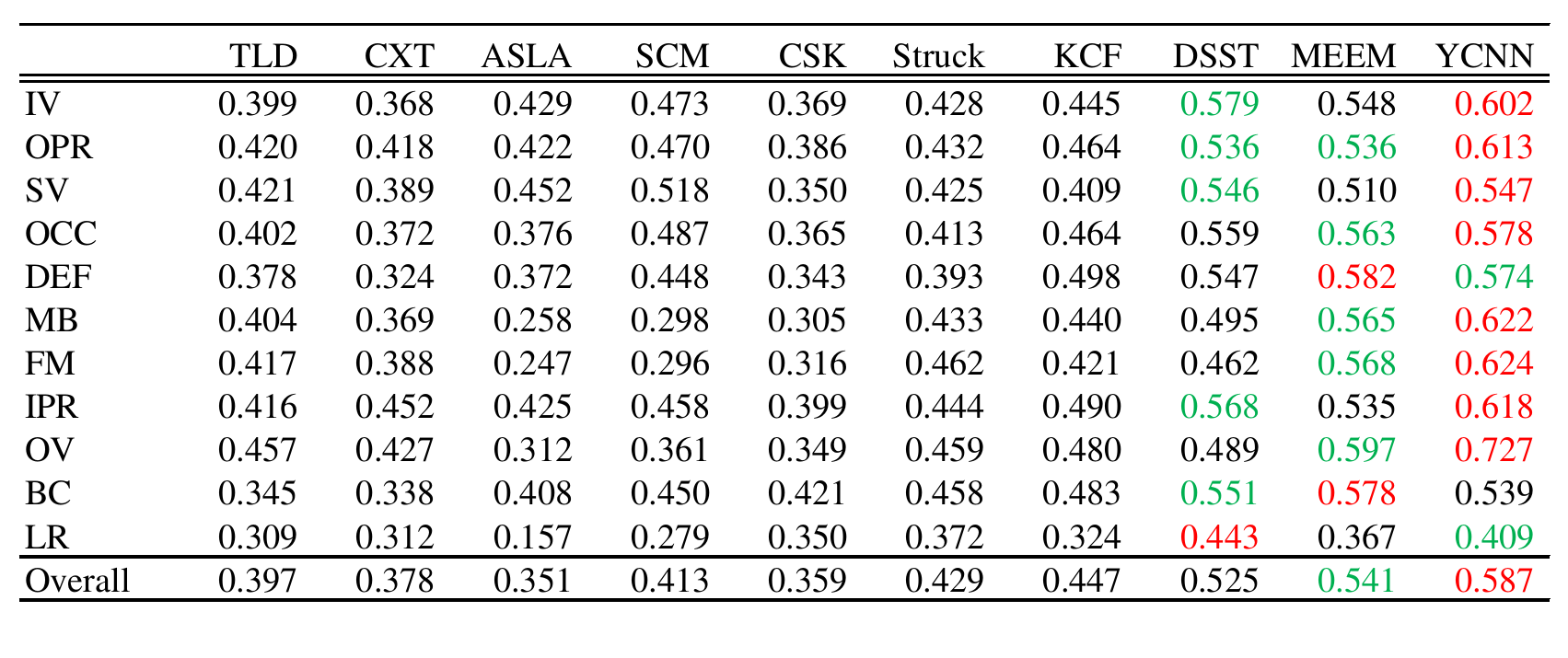}
\end{table}

\subsection{Attribute Based Comparison}

A typical tracking sequence may contain a variety of challenges, such as illumination variation (IV), out-of-plain rotation (OPR), scale variation (SV), occlusion (OCC), deformation (DEF), motion blur (MB), fast motion (FM), in-plain rotation (IPR), out-of-view (OV), background cluttered (BC), and low resolution (LR). To analysis the ability of handling different challenges, the tracking results are further evaluated on those sequences with the 11 different attributes.

\begin{figure}[t]
  \centering
  \includegraphics[width=1.0\linewidth]{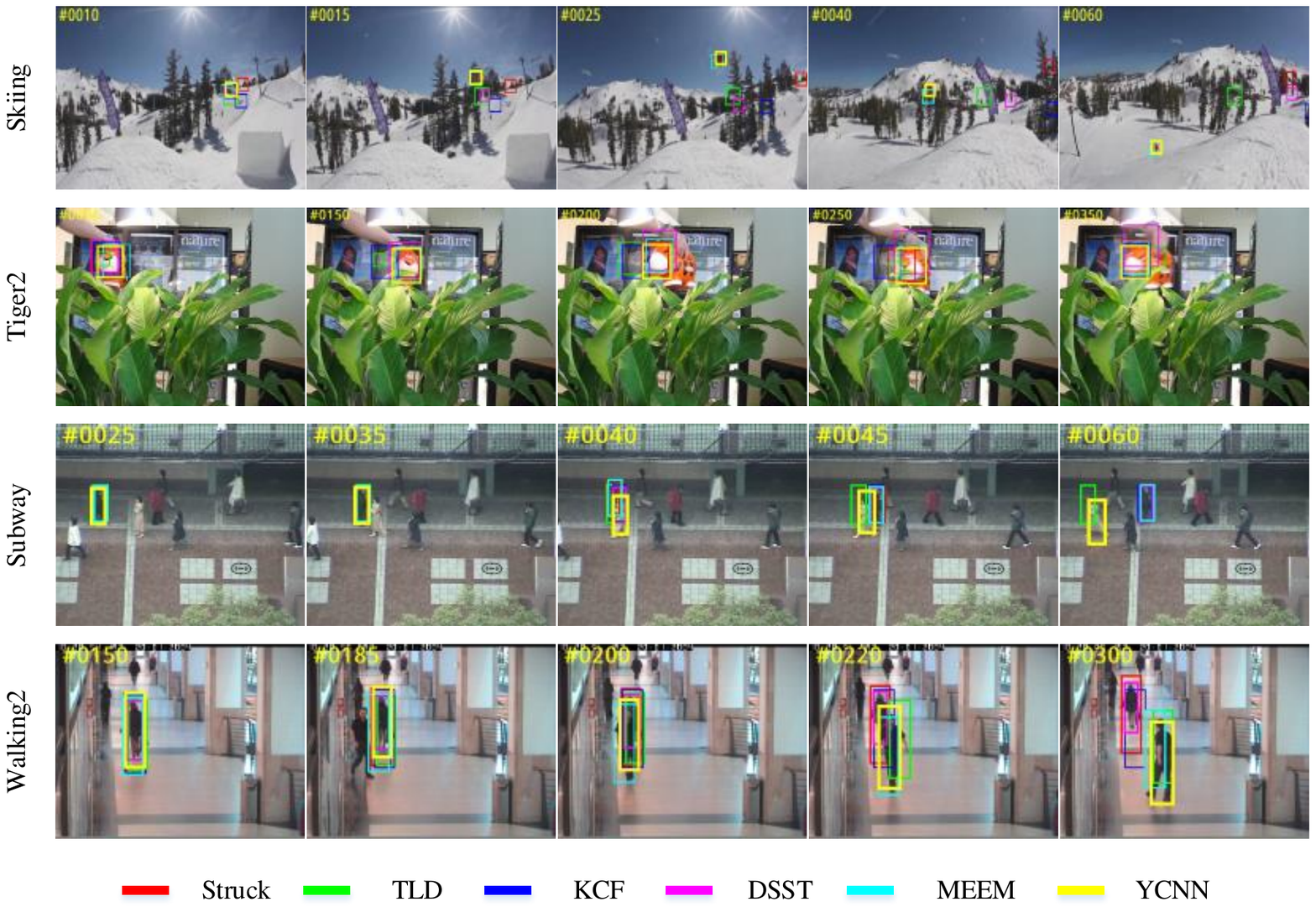}
  \caption{Qualitative comparisons between Struck, TLD, KCF, DSST, MEEM, and YCNN. The YCNN works well in \textit{Skiing} and \textit{Tiger2} but fails in \textit{Subway} and \textit{Walking2}.}
  \label{fig:qualitative_comparison}
\end{figure}

The results based on center location error and overlap rate are shown in table \ref{tab:precision_attribute_analysis} and table \ref{tab:sucess_attribute_analysis} respectively. The results have shown that, the proposed algorithm achieves an outstanding performance in most of the attribute based comparisons, especially when there exists high contrast between the object and the background. As shown in figure~\ref{fig:qualitative_comparison}, the targets in sequence \textit{Skiing} and \textit{Tiger2} are of high saliency and tracked by YCNN accurately. But it does not work that well when handling background-cluttered sequences. For example, in sequence \textit{subway} and \textit{Walking2}, the YCNN all drifts from the true object when a similar object appears in the search area. This problem may be alleviated with more training sequences.

\begin{table}[t]
  \caption{Implementation details and tracking speed comparisons.}
  \label{tab:speed_comparison}
  \centering
  \begin{tabular}[t]{l r r r r}
    \hline
     & framework& GPU & language & fps \\
     \hline \hline
     H. Li \textit{et al.}~\cite{DeepTrack}& CUDA-PTX & GTX 770 & Matlab & 1.3\\
     L. Wang \textit{et al.}~\cite{FCN_GS} & Caffe & GTX TITAN & Matlab & 3 \\
     H. Nam \textit{et al.}~\cite{MDCNN} & MatConvNet & Tesla K20m & Matlab & 1 \\
     Ours (YCNN)  & TensorFlow & Tesla K40c & Python & 45\\
     \hline
  \end{tabular}
\end{table}

\subsection{Speed Analysis}

A good tracker should not only track an object accurately but also run fast. The traditional CNN-based tracking algorithms, though have achieved great success in terms of accuracy and robustness, but are also charged with low speed. We have listed the implementation details and tracking speed of some recent CNN-based tracking algorithms and our proposed YCNN in table~\ref{tab:speed_comparison}. The trackers proposed in~\cite{DeepTrack,FCN_GS,MDCNN} all runs slowly, which is mainly due to the frequent retraining and updating of the CNN. However, our proposed YCNN runs at a very high speed of 45 frames-per-second, which is several times the other CNN-based trackers regardless of the differences in implementation details. This is because no backpropagation is needed in the two-flow CNN when tracking.

\section{Conclusion}

We have proposed a novel two-flow CNN for visual tracking in a more generic way. The YCNN reformulates the tracking problem as similarity measurement between object and search candidates. Once the YCNN is properly trained, it can be used to track all kinds of object. The experiments have shown that, our proposed YCNN can achieve an outstanding performance while run at high speed.





\bibliographystyle{plainnat}
\bibliography{once_for_all}







\end{document}